\documentclass[fleqn,10pt]{wlscirep}
\usepackage[utf8]{inputenc}
\usepackage[T1]{fontenc}
\usepackage{acronym}
\newacro{SA}{Simulated annealing}
\newacro{CIM}{coherent Ising machine}
\newacro{SDP}{semi-definite programming}
\newacro{QA}{quantum annealing}
\newacro{CITS}{coherent Ising tree search}
\newacro{MCTS}{Monte Carlo tree search}
\newacro{VMM}{vector-matrix-multiplication}
\newacro{FPGA}{field-programmable gate array}

\title{CITS: Coherent Ising Tree Search Algorithm Towards Solving Combinatorial Optimization Problems}

\author[1]{Yunuo Cen}
\author[1]{Debasis Das}
\author[1,*]{Xuanyao Fong}
\affil[1]{Department of Electrical and Computer Engineering, National University of Singapore, Singapore 117583}

\affil[*]{kelvin.xy.fong@nus.edu.sg}

\begin{abstract}
\ac{SA} attracts more attention among classical heuristic algorithms because the solution of the combinatorial optimization problem can be naturally mapped to the ground state of the Ising Hamiltonian. 
However, in practical implementation, the annealing process cannot be arbitrarily slow and hence, it may deviate from the expected stationary Boltzmann distribution and become trapped in a local energy minimum. 
To overcome this problem, this paper proposes a heuristic search algorithm by expanding search space from a Markov chain to a recursive depth limited tree based on \ac{SA}, where the parent and child nodes represent the current and future spin states. 
At each iteration, the algorithm will select the best near-optimal solution within the feasible search space by exploring along the tree in the sense of `look ahead'.
Furthermore, motivated by \ac{CIM}, we relax the discrete representation of spin states to continuous representation with a regularization term and utilize the reduced dynamics of the oscillators to explore the surrounding neighborhood of the selected tree nodes.
We tested our algorithm on a representative NP-hard problem (MAX-CUT) to illustrate the effectiveness of this algorithm compared to \ac{SDP}, \ac{SA}, and simulated \ac{CIM}.
Our results show that above the primal heuristics \ac{SA} and \ac{CIM}, our high-level tree search strategy is able to provide solutions within fewer epochs for Ising formulated NP-optimization problems.
\end{abstract}
\begin{document}

\flushbottom
\maketitle
\thispagestyle{empty}
\section*{Introduction}

Many combinatorial optimization problems (\emph{e.g.}, VLSI floorplanning\cite{anand2012customized}, drug discovery\cite{matsubara2020digital}, and advertisement allocation\cite{tanahashi2019application}) aim to find the optimal solution among a finite set of feasible solutions. 
However, finding the exact solution of combinatorial optimization problems, generally, requires the exploration of the entire solution space, which increases exponentially in size with the size of the optimization problem and making it intractable to solve exactly\cite{lucas2014ising}. 
As a result, the research community has significant interest in algorithms to find near-optimal solutions within a reasonable time.

\begin{figure}[ht]
\centering
\includegraphics[width=\linewidth]{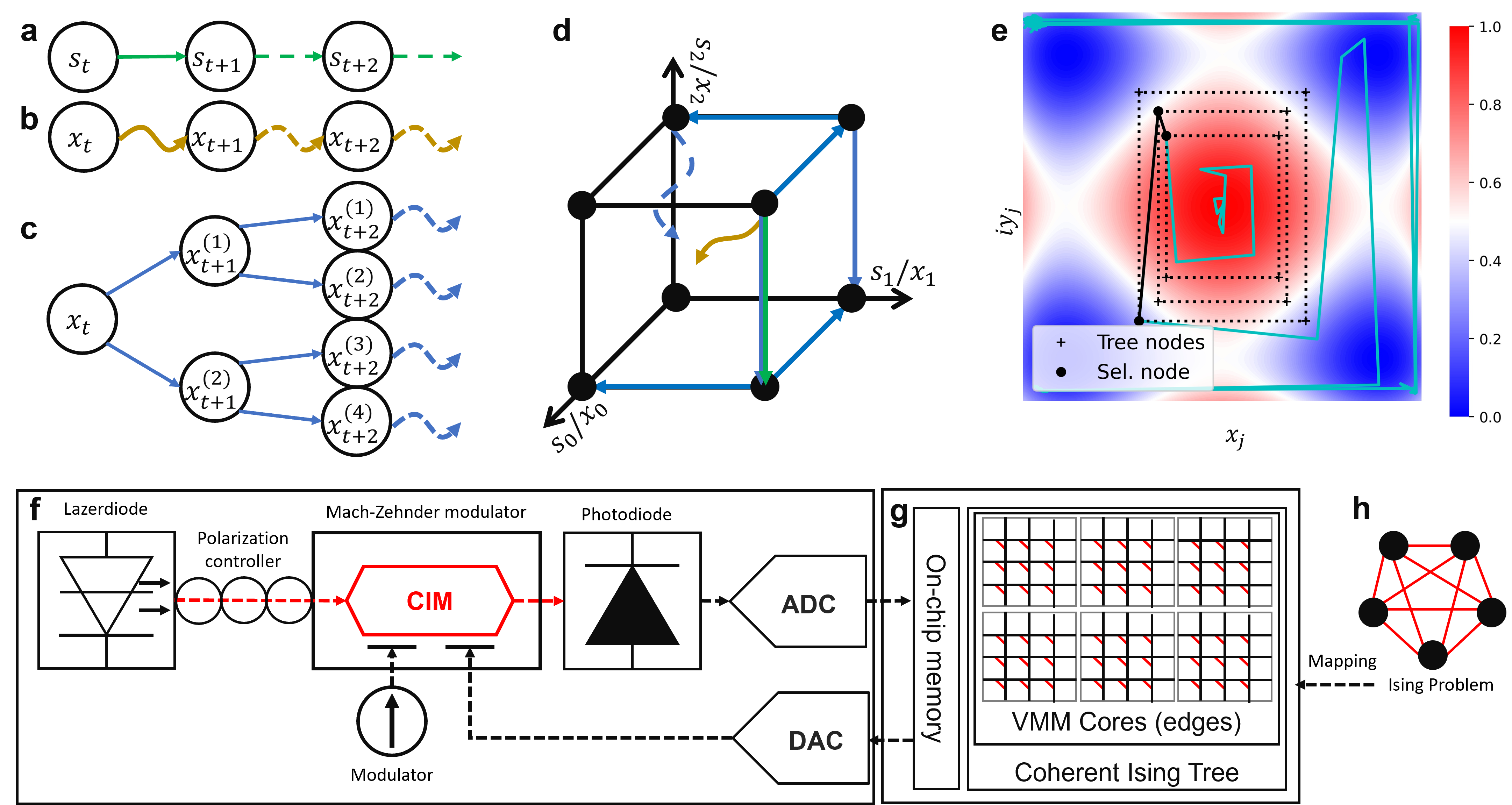}
\caption{\textbf{Search space of \ac{SA}, \ac{CIM} and \ac{CITS} within three coupling Ising spins:} \textbf{(a)} The Markov chain of \ac{SA}, where the straight solid/dot lines represent the Metropolis-Hasting sampling at current/future time step.
\textbf{(b)} The Markov chain of \ac{CIM}, where the curve solid/dot lines represent the oscillator dynamics at current/future time step.
\textbf{(c)}. The Ising tree structure of \ac{CITS}, where the straight lines represent the expansion based on \ac{SA} and the curve line represents the exploration based on \ac{CIM}; 
\textbf{(d)} Intuitive explanation of annealing mechanism of \ac{CITS}. From the initial spin configuration with high Ising Hamiltonian, expanding the search space by primal heuristic \ac{SA} (blue straight lines), then exploring the two local space in parallel by primal heuristic \ac{CIM} (blue curve lines).
\textbf{(e)} Explanatory annealing process on two uncoupling Ising spin. Each dashed triangle represent the child nodes of the corresponding time step. Only the most potential node is selected for future time steps. 
\textbf{(f)} Experimental schematic of poor man's \ac{CIM}\cite{bohm2019poor}, where we abstract and simulate the reduced dynamics since we only care about the solution quality instead of measuring the whole system. 
\textbf{(g)} Hardware design of \ac{CITS} interfaces with poor man's \ac{CIM}. The \ac{VMM} cores compute the Hamiltonian and the reward of the child nodes. This part can be paralleled and accelerated using FPGA or non-volatile memory technology. The on-chip memory stores the Ising spin configurations of each nodes, and the corresponding Hamiltonian. 
\textbf{(h)} Ising formulation of combinatorial optimization problem that can map the edge values to the cells in \ac{VMM} cores.}
\label{fig:seach_space}
\end{figure}

Among the approaches in the literature, the Ising model (Fig.~\ref{fig:seach_space}(h)) has attracted the most attention because it is straightforward to map from many combinatorial optimization problems\cite{lucas2014ising}. 
Moreover, two well-known heuristic algorithms have been studied extensively: \ac{QA}\cite{johnson2011quantum} and its classical counterpart, \ac{SA}\cite{kirkpatrick1983optimization}. 
In these algorithms, the cost function is directly encoded as the Ising Hamiltonian and feasible solutions to the combinatorial optimization problems are obtained by searching for the energy minima. 
Furthermore, by the adiabatic theorem of quantum mechanics or the Boltzmann distribution of statistical mechanics\cite{mitra1986convergence}, \ac{QA} and \ac{SA} can obtain the exact solution if the annealing time is large enough. 
However, the \ac{QA} is usually implemented on superconducting qubits, which can be costly.
However, the Ising spins for \ac{SA} can be hardware-friendly for conventional computers\cite{mu202120x28}, especially for CMOS-Compatible spintronics implementation\cite{sutton2017intrinsic,shim2017stochastic,mondal2020ising,andrawis2020antiferroelectric}. 
Regardless of implementation, to achieve the Boltzmann distribution using the annealing process may become intractable. 
By only considering the flipping probabilities of identical spins to achieve a quasi-equilibrium distribution is more practical but usually traps the \ac{SA} in a local optimum\cite{hibat2021variational}. 
Some studies try to avoid this situation by introducing noise\cite{mondal2020ising} but the effectiveness is still under debate.

Recently, the \ac{CIM} has garnered much research interest because its bistable coherent states can be naturally mapped to the Ising Hamiltonian\cite{marandi2014network,mcmahon2016fully,bohm2019poor}. 
Compared to \ac{SA}, \ac{CIM} shows theoretical advantage of facilitating a quantum parallel search across multiple regions. 
Compared to \ac{QA}, \ac{CIM} shows physical advantage of room temperature operating conditions and CMOS-compatibility, which benefit from the fact that the information loss due to the measurement-feedback scheme for capturing the reduced dynamics of \ac{CIM} seems to be unimportant\cite{mcmahon2016fully}. 
Moreover, the full potential of \ac{CIM} has yet to be explored.

To overcome the aforementioned challenges, we propose a tree search algorithm, which we call as \ac{CITS}, that combines two primal heuristic (\ac{SA} and \ac{CIM}) to find near-optimal solutions.
The proposed algorithm is inspired by \ac{MCTS}, a best-first search algorithm that expands the search tree based on random exploration and eventually taking the most promising move\cite{coulom2006efficient}. 
CITES is a recursive size limited search algorithm following the idea of `look-ahead'. 
The main differences with \ac{MCTS} are as follows:
\begin{itemize}
    \item \ac{MCTS} expands child nodes for all possible moves at each time step, and the overall tree is needed for the future time steps. 
    The expansion step of \ac{CITS} is as follows: the child node corresponding to the selected spin state becomes a root node, and the unselected nodes are pruned away, eventually expanding the child nodes to a tree (limited to predefined depth and breadth) based on the primal \ac{SA} heuristic. 
    At each time step, \ac{CITS} generates a size limited tree and hence, it is a recursive size limited search algorithm.
    \item \ac{MCTS} only visits one path in the tree from the root to the leaf node at each time step while \ac{CITS} searches the whole Ising tree;
    \item \ac{MCTS} is usually applied on zero-sum and complete information games so each leaf node has an exact reward value. \ac{CITS} is applied on combinatorial optimization optimization problems in which the best optimum is not known. Thus, the change in Ising Hamiltonian determines the leaf value;
    \item \ac{MCTS} predicts leaf value by performing playout with random moves to end the game whereas \ac{CITS} evaluates the Hamiltonian of the Ising spin configurations of all child nodes and the rewards of child nodes are backpropagated to their parent nodes;
\end{itemize}

In \ac{CITS}, a Markov chain in conventional \ac{SA} or \ac{CIM} is expanded to an `Ising tree', where the future spin states are also taken into consideration (Fig.~\ref{fig:seach_space}a-c).
Instead of simply searching for a lower Ising Hamiltonian in Stone space $\left(\{-1,+1\}^n\right)$ using \ac{SA} or in real coordinate space $\left(\mathbb{R}^n\right)$ based on the reduced dynamics of \ac{CIM}, the high-level strategy of \ac{CITS} leverages both primal \ac{SA} and \ac{CIM} heuristics in both search spaces to obtain solutions of the Ising formulated problems (Fig.~\ref{fig:seach_space}d). 
The dashed rectangles in Fig.~\ref{fig:seach_space}e describes the depth and the breadth of our \ac{CITS} algorithms.
At each time step, only the most prominent node is selected for the future time step.
Particularly, the computation speed when utilizing the measurement-feedback scheme is mainly limited by the communication time between analog (Fig.~\ref{fig:seach_space}f) and digital domain (Fig.~\ref{fig:seach_space}g).
As a result, linearly scaling up (tree depth $d$ and breadth $b$) the computation in the digital domain will require more computational resource, especially the amount of \ac{VMM} cores and the size of memory. 
However, the total time of each cycle will not increase too much as long as the memory can be realized on-chip, such that the computation time, \emph{i.e.} on FPGA, can be orders of magnitude smaller than the communication time.
In this paper, we demonstrate that \ac{CITS} tradeoff an acceptable spatial complexity to accelerate the search to find better solutions on MAX-CUT problems (as compared to \ac{SA} or \ac{CIM}) with graph sizes ranging from 36 nodes to 230 nodes. 

\section*{Results}

\subsection*{Reduced dynamics of coherent Ising tree in Lagrange picture}
Ising model (Fig~\ref{fig:seach_space}c)) describes the behaviour of Ising spins and interactions between each other. The general form of Ising Hamiltonian intriguing classical heuristic algorithms is shown as\cite{sherrington1975solvable}:

\begin{equation}
H\left(s_1,s_2,...,s_n\right)=-\frac{1}{2}\sum_{j}^{n}\sum_{l}^{n}J_{jl}s_{j}s_{l}
\label{eq:classical_H}
\end{equation}
$s_j$ is the state of the $j^{th}$ spin in Ising model, which only has two states (spin-up or spin-down) represented by binary $+1$ or $-1$. 
$J_{jl} \left(=-J_{jl}\right)$ is the coupling coefficient between the $j^{th}$ and the $l^{th}$ spins, and represents ferromagnetic (antiferromagnetic) coupling is it is positive (negative)\cite{ising1925beitrag}. 
For any Ising formulated combinatorial optimization problem, the aim is to encode the problem with an Ising Hamiltonian and explore the search space to find a spin configuration that minimizes the Hamiltonian\cite{lucas2014ising}.
Except for binary representation, the Ising spins can be modelled as wave functions and used the phase difference to represent spin-up ($0$-phase) or spin-down ($\pi$-phase).
Note that additional constraints is required for the phase degeneracy of Ising spins when mapping from combinatorial optimization problems, \emph{i.e.} second harmonic injection locking\cite{wang2019oim} for classical approach or down conversion\cite{wang2013coherent} for quantum approach. 
In our approach, we start from quantum harmonic oscillators $\hbar\omega_0\hat{a}^\dagger\hat{a}$, and a parametric nonlinear (trigonometric) feedback signal from the external field is utilized to degenerate and couple the Ising spins:

\begin{equation}
\hat{H}\left(\hat{a}_1,\hat{a}_2,...,\hat{a}_n\right)= \sum_j^n\left[\hbar\omega_0\hat{a}_j^\dagger\hat{a}_j+
\hbar\omega_\alpha\cos\left(\frac{\omega_0}{2\omega_\alpha}\left(\hat{a}_j^\dagger+\hat{a}_j\right)+
\frac{\omega_0}{2\omega_\beta}\sum_l^nJ_{jl}\left(\hat{a}_l^\dagger+\hat{a}_l\right)
\right)+\Xi
\right]
\label{eq:quantum_H}
\end{equation}
where $\left(\hat{a}^\dagger, \hat{a}\right)$ are the creation and annihilation operator, $\hbar$ is the Dirac constant, $\omega_0$ is the intrinsic frequency of the Ising spins. $\omega_\alpha$ denotes the frequency of the injected feedback signals that pump the Ising spins into two coherent states, $\omega_\beta$ denotes the frequency of the mutual coupling signals encoding the Ising Hamiltonian as shown in Eq.~\eqref{eq:classical_H}. $\Xi$ is the random diffusion term modelled as 0-mean Gaussian white noise. 

\begin{figure}[ht]
\centering
\includegraphics[width=\linewidth]{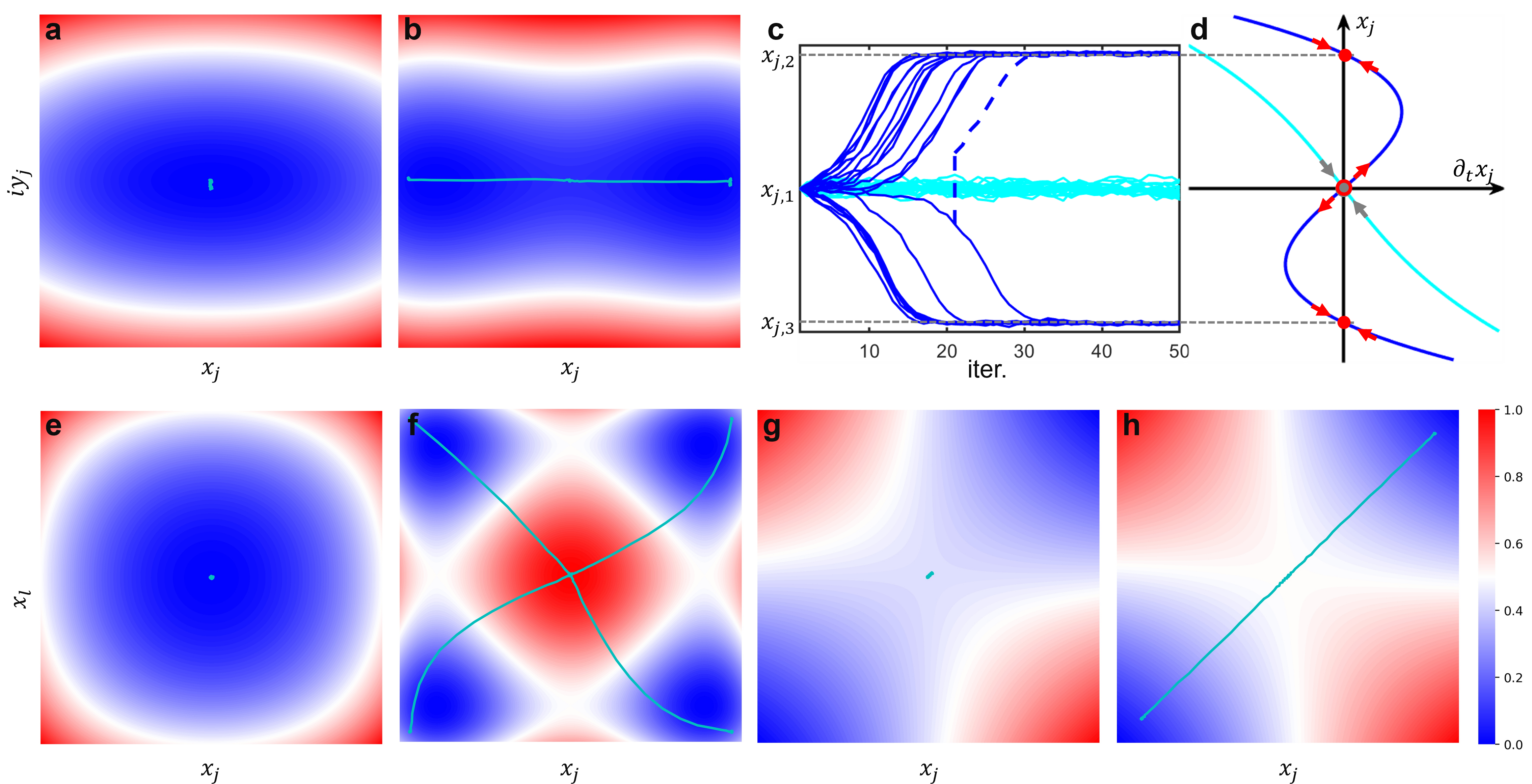}
\caption{\textbf{Reduced dynamics of parametric nonlinear (trigonometric) oscillator:} 
\textbf{(a-b)} Energy landscape and trajectory of uncoupled Ising spins when increasing the feedback gain from  below threshold ($\alpha=0.8$) to above threshold ($\alpha=1.3$).
\textbf{(c)} Time evolution of the real part of the reduced dynamics with feedback gain $\alpha=0.8$ (cyan) and $\alpha=1.3$ (blue), where the dash line indicates flipping the sign of the spins will not affect the bistability. 
\textbf{(d)} Stability analysis of the uncoupled spins. When the feedback gain is below threshold, the real part of the reduced dynamics will only have one stable fixed points at $x_1 = 0$ (gray dot and arrow). When the feedback gain is above threshold, the stable fixed points at $x_1 = 0$ become unstable (red ring and arrow), and there exists two symmetric stable fixed points at $x_2 = -x_3$ (red dot and arrow). 
\textbf{(e-f)} Projected energy landscape on real axis and trajectory of two uncoupled Ising spins when increasing the feedback gain from below threshold ($\alpha=0.8$) to above threshold ($\alpha=1.3$). 
\textbf{(g-h)} Projected energy landscape on real axis and trajectory of two coupled Ising spins when increasing the feedback gain from below threshold ($\alpha=\beta=0.5$) to above threshold ($\alpha=\beta=0.6$). }
\label{fig:reduced_dynamic}
\end{figure}

The quantum-inspired algorithms excavate the potential toward solving combinatorial optimization problems by leveraging the underlying reduced dynamics of Hamiltonian systems. 
Based on Heisenberg equation, the motion of the annihilation operators can potentially describe an optimization pathway toward the global optimum. 
Cooperating with Langevin equation, the Hamiltonian in Eq. (2) can be translate into Heisenberg picture\cite{wang2013coherent}. 
A classical approach is to approximate the expectation of the Hamiltonian by $\langle\Phi|\hat{H}|\Phi\rangle$ using complex representation $\Phi=[\phi_j]=[x_j+iy_j]$\cite{goto2019combinatorial}.
And Lagrangian is able to capture the classical approximation of the reduced dynamics of two quadrature components.
As a result, the reduced dynamics of the Ising spins in Lagrange picture becomes:

\begin{equation}
\begin{aligned}
\frac{\partial x_j}{\partial t}
&=-\lambda_x\frac{\partial\langle\Phi|\hat{H}|\Phi\rangle}{\partial x_j}\
=\hbar\omega_0\sin\left(2\alpha x_j+2\beta\sum_{l=1}^nJ_{jl}x_l\right) -2\hbar\omega_0x_j+2\hbar\omega_0\frac{d\xi_{x,j}}{dt}\\
\frac{\partial y_j}{\partial t}
&=-\lambda_y\frac{\partial\langle\Phi|\hat{H}|\Phi\rangle}{\partial y_j}
=-2\hbar\omega_0y_j+2\hbar\omega_0\frac{d\xi_{y,j}}{dt}
\label{eq:reduced_dynamic}
\end{aligned}
\end{equation}
where $\lambda_x, \lambda_y(=1)$ are the Lagrange multiplier of $x$ and $y$ component.
$\alpha=(\omega_0/2\omega_\alpha)$ is the feedback gain and $\beta=(\omega_0/2\omega_\beta)$ is the coupling gain. As we discretize Eq.~\ref{eq:reduced_dynamic} using Euler method with $\Delta{t}=1/2\hbar\omega_0$, the reduced dynamics are mathematically equivalent to the poor man's \ac{CIM}\cite{bohm2019poor} if we neglect the Gaussian white noise. Note that, in the rest of this paper, we use a discrete version to model the reduced dynamics in sense of measurement-feedback approach.

\begin{equation}
\begin{aligned}
x_j[t+1] &= x_j[t] + \frac{\partial x_j[t]}{\partial t}\Delta{t} 
=\frac{1}{2}\sin\left(2\alpha x_j[t]+2\beta\sum_{l=1}^nJ_{jl}x_l[t]\right)+\xi_{x,j}[t]
\\
y_j[t+1] &= y_j[t] + \frac{\partial y_j[t]}{\partial t}\Delta{t} = \xi_{y,j}[t]
\label{eq:discrete_dynamic}
\end{aligned}
\end{equation}
$x_j[t],y_j[t]$ are the pair of canonical conjugate variables for the $j^{th}$ Ising spin with time derivatives of $\partial x_j[t]/\partial t,\partial y_j[t]/\partial t$ at the $t^{th}$ time step, where many uncoupled spins are simulated for 50 time steps and the results are shown in Fig.~\ref{fig:reduced_dynamic}a-d.
Fig.~\ref{fig:reduced_dynamic}a,b show the representative trajectories under different energy landscapes with different feedback gain.
When the feedback gain is below the threshold, the landscape of the Hamiltonian has only one energy minimum, and the Ising spins are squeezed within the vacuum state. 
As the feedback gain increases above the threshold, the landscape of the Hamiltonian will have two energy minima.
Eventually, the Ising spins will bifurcate into two coherent states, corresponding to spin-up and spin-down. 
The real part of the reduced dynamics of 50 Ising spins are shown in Fig.~\ref{fig:reduced_dynamic}c, where $x_{j,1}=0$ corresponds the vacuum state and $x_{j,2}=-x_{j,3}$ correspond to the coherent states.
The implication is described in Fig.~\ref{fig:reduced_dynamic}d. 
When the feedback gain is below there threshold, the stable fixed points are at $x_{j,1}=0$.
When the feedback gain is above threshold, the fixed points at $x_{j,1}=0$ become unstable whereas symmetric stable fixed points appear at $x_{j,2}=x_{j,3}$. 
Furthermore, from Eq.~\eqref{eq:discrete_dynamic}, we observe that the imaginary part of the reduced dynamics remains around 0 under perturbation of 0-mean Gaussian white noise. 
Fig.~\ref{fig:reduced_dynamic}e,f/g,h also show the energy landscapes and representative trajectories of the real part of the dynamics of uncoupled/coupled Ising spins.
Similarly, when the gains are below/above threshold, Ising spins enter the vacuum/coherent states depending on whether nor not they are coupled.

The evolution and expansion steps (see Method Section) in our \ac{CITS} algorithm is based two primal heuristics, \ac{CIM} and \ac{SA}. 
The intuition of cooperating these two algorithms is from the symmetric energy landscapes and trajectories shown in Fig.~\ref{fig:reduced_dynamic}.
The blue dash line reveals that, flipping the sign of the Ising spin will not affect the bistability of the reduced dynamics (Eq.~\eqref{eq:reduced_dynamic}, which are known as odd functions).
During the evolution and expansion step, \ac{CITS} can explore along real coordinate space based on Eq.~\eqref{eq:discrete_dynamic} while expand the search space along Stone space based on Eq.~\eqref{eq:prob}. 
As a result, \ac{CITS} is able to boost \ac{CIM} toward exploring multiple search spaces in the future time steps without losing the bistability.
Fig.~\ref{fig:seach_space}c and dash lines in Fig.~\ref{fig:seach_space}e shows the coherent Ising tree with depth $d=2$ and breadth $b=2$ at time step $t$. Compared to four identical and independent trajectories in Fig.~\ref{fig:reduced_dynamic}f, the search space coverage of a coherent Ising tree is even broader, which gives an intuition that \ac{CITS} can help escape trapping in local minima.

\subsection*{Ablation study: complete and naive search schemes}

\begin{figure}[ht]
\centering
\includegraphics[width=\linewidth]{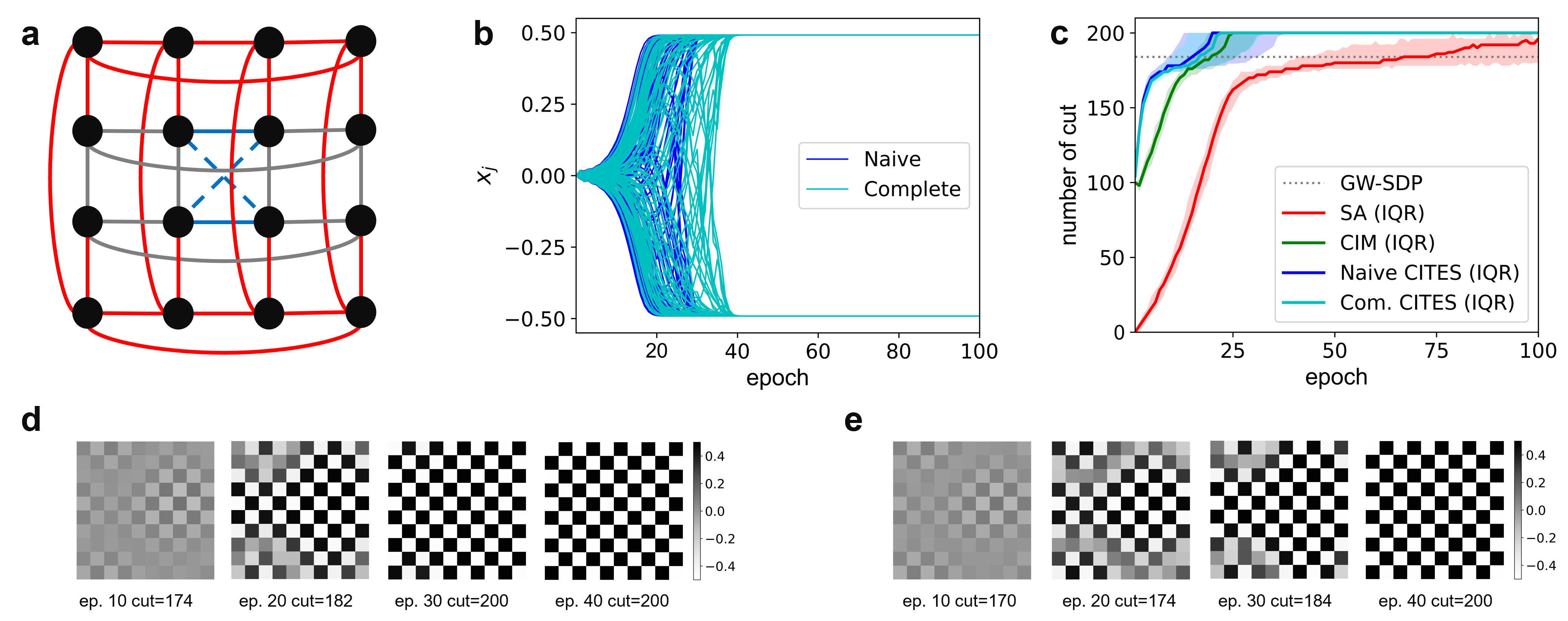}
\caption{\textbf{Benchmark results on square lattice graphs:} 
\textbf{(a)} Graph structure of grid lattice with two periodic dimensions (all solid lines). Circular ladder can be obtained by only choosing the vertices connected by the edges shown in blue and gray solid line. And Mobius ladder can be obtained by twisting the blue solid lines to blue dash lines in circular ladder graphs. 
\textbf{(b)} A simulation of time evolution of spin amplitude on a 10-by-10 square lattice using two schemes (with feedback strength $\alpha=0.25$, and coupling strength $\beta=0.29$). 
\textbf{(c)} The number of cut on 10-by-10 square lattice given by \ac{SA}, \ac{CIM}, \ac{CITS} for 100 epochs.
\textbf{(d)} The Ising spin configuration at epoch 10, 20, 30 and 40, where two end of the color bar represent two stable solution of the oscillator (black for spin-up and white for spin-down) for naive scheme and \textbf{e)} complete scheme.}
\label{fig:10by10lattice}
\end{figure}

To evaluate the efficiency of \ac{CITS} as compared to \ac{SA} and \ac{CIM}, we run the simulations for 100 epochs for each instance. 
First, we evaluate on a 10-by-10 square lattice with two periodic dimensions.
In the Method section, we proposed two expansion (and exploration) schemes for \ac{CITS}. 
The naive scheme only considers the search spaces (tree nodes) generated by the primal heuristic, \ac{SA}.
The complete scheme explores around the tree nodes based on primal heuristic, \ac{CIM}.
In Fig.~\ref{fig:10by10lattice}b, we show the time evolution of amplitudes corresponding to 10-by-10 square lattice graph (Fig.~\ref{fig:10by10lattice}a) of the two exploration schemes.
In our simulation, the initialization could be essential for the rest of the epochs, just like the weight initialization for neural networks.
The impact of initialization are beyond the scope of the proposed tree search algorithms and are left out of this article.
After initialization around the unstable fixed point $x_1=0$, \ac{CITS} explores the search space and eventually stabilizes at two symmetric fixed points $x_2=-x_3$ representing spin-up and spin-down. 
Fig.~\ref{fig:10by10lattice}d,e shows the explicit spin configurations at epoch 10, 20, 30, 40 corresponding to the time evolution of the spin amplitudes.
In the first few epochs, the spin configurations form an organized region at the middle right of the square lattice. 
After around 10 epochs, the amplitudes of Ising spins bifurcate, which give rise to the coupling effect.
In the naive scheme, the spin configuration at epoch 20 is sub-optimal.
At epoch 30, the system stabilizes at the global energy minimum for the rest of the time step.
In the complete scheme, the convergence speed is slower but it eventually reaches the ground state at epoch 40.

Though we cannot claim which scheme is better from only one simulation result, the visualization of the underlying dynamics can help us gain some intuition on the optimization process.
100 simulations were run for both schemes to evaluate which one can have a faster convergence speed.
Fig.~\ref{fig:10by10lattice}c shows the number of cuts on the 10-by-10 square lattice given by the solution at each epoch.
To further evaluate the performance of \ac{CITS}, we benchmark this result with SDP, \ac{SA} and simulated \ac{CIM} with the setup mentioned in the Methods Section.
Since we are comparing the performance between annealing-based heuristic algorithms, the approximate solution of 184 given by SDP is used as a baseline to evaluate the epochs-to-solution and the success rate of proposed \ac{CITS} algorithm and its two primal heuristics, \ac{SA} and simulated \ac{CIM}.
We run each these three algorithms 100 times with different random seed numbers.
Since the performance of each run varies, we present the interquartile range (IQR, range of between median of upper half $Q_{75}$ and median of lower half $Q_{25}$) on Fig.~\ref{fig:10by10lattice}c, where the solid lines are the median values $Q_{50}$.
In the beginning, \ac{CITS} and \ac{CIM} are initialized at vacuum state, and suddenly obtain a 0.5-approximation solution due to the diffusion term introduced by the Gaussian white noise.
Notably, during the first 7 epochs, the increment in number of cut for \ac{CITS} and \ac{CIM} are around 10 and 5 per epoch, respectively. 
\ac{CITS} has a faster convergence to an approximate solution because deeper nodes in the coherent Ising tree tend to explore the tree in the coming future time step and based on the energy change, \ac{CITS} will decide the best move among the search space.
In the beginning of the annealing process, the spin configurations are chaotic and a random flip tends to lower the Ising energy.
Consequently, \ac{CITS} tends to accept the further time steps.
For the complete scheme, fine-grained local search on each node is performed.
As \ac{CITS} goes from root node to one of the leaf node, its primal heuristic \ac{CIM} needs to run for $d$ times within a epoch.
Since the tree depth is 2, \ac{CITS} achieves approximately 2x faster convergence to the lower Hamiltonian as compared to \ac{CIM}.
Notably, the naive scheme only performs coarse-grained local search on each node, where the primal heuristic \ac{CIM} is not run to explore the surrounding search space.
Intuitively, the fine-grained local search on each node provides more information of the Ising system, \emph{i.e.} the reduced dynamics.
However, the naive scheme seems to converge faster than the complete scheme.
This indicates that instead of fine-grained local search spaces, the breadth and the depth of tree are the key contributors to the performance of \ac{CITS}. 
Since the naive scheme saves a lot computational complexity, the simulation of \ac{CITS} is based on naive scheme in the rest of this paper.

After a few epochs, the landscape of Ising Hamiltonian becomes more complicated.
In this scenario, both algorithms tend to be trapped  in the local minima and slow down the annealing process, which takes 11/18/27 epochs (for lower/median/upper quartile) for \ac{CITS} to outperform SDP and 15/23/33 epochs to reach the global optimum.
In comparison, it takes 15/20/26 epochs for \ac{CIM} to outperform SDP and 19/25/33 epochs to reach the global optimum, respectively.
The speed of convergence of \ac{CITS} depends not only depends the `depth' of the coherent Ising tree but also the `breadth'.
A wider coherent Ising tree provides a larger search space, which potentially contains a spin configuration with a lower Hamiltonian.
As mentioned in the Introduction section, the expansion of the coherent Ising tree is based on another primal heuristic \ac{SA}, which is able to identify the most promising search spaces based on the highest flipping probability computed by Eq.~\eqref{eq:prob}. 
Hence, we also compared \ac{CITS} with \ac{SA}, where $Q_{25}$ needs 79 epochs to reach the ground state.
Note that the algorithms do not always outperform SDP within 100 epochs, where it takes 41/67 epochs for $Q_{25}/Q_{50}$. 
The Ising spins of \ac{SA} do not have the properties of the `vacuum state' since \ac{SA} initializes spin configuration with all spin-up (or all spin-down) with 0 cut in the beginning.
The number of spins allowed to be flipped is limited so as to preserve the quasi-equilibrium distribution given by the approximation and ensure that the Markov process will converge to a stable distribution.
Nonetheless, it can outperform SDP and may possibly return a near-optimal solution in a longer but acceptable time scale.

\subsection*{Benchmarking \ac{CITS} on different MAX-CUT instances}

\begin{figure}[ht]
\centering
\includegraphics[width=\linewidth]{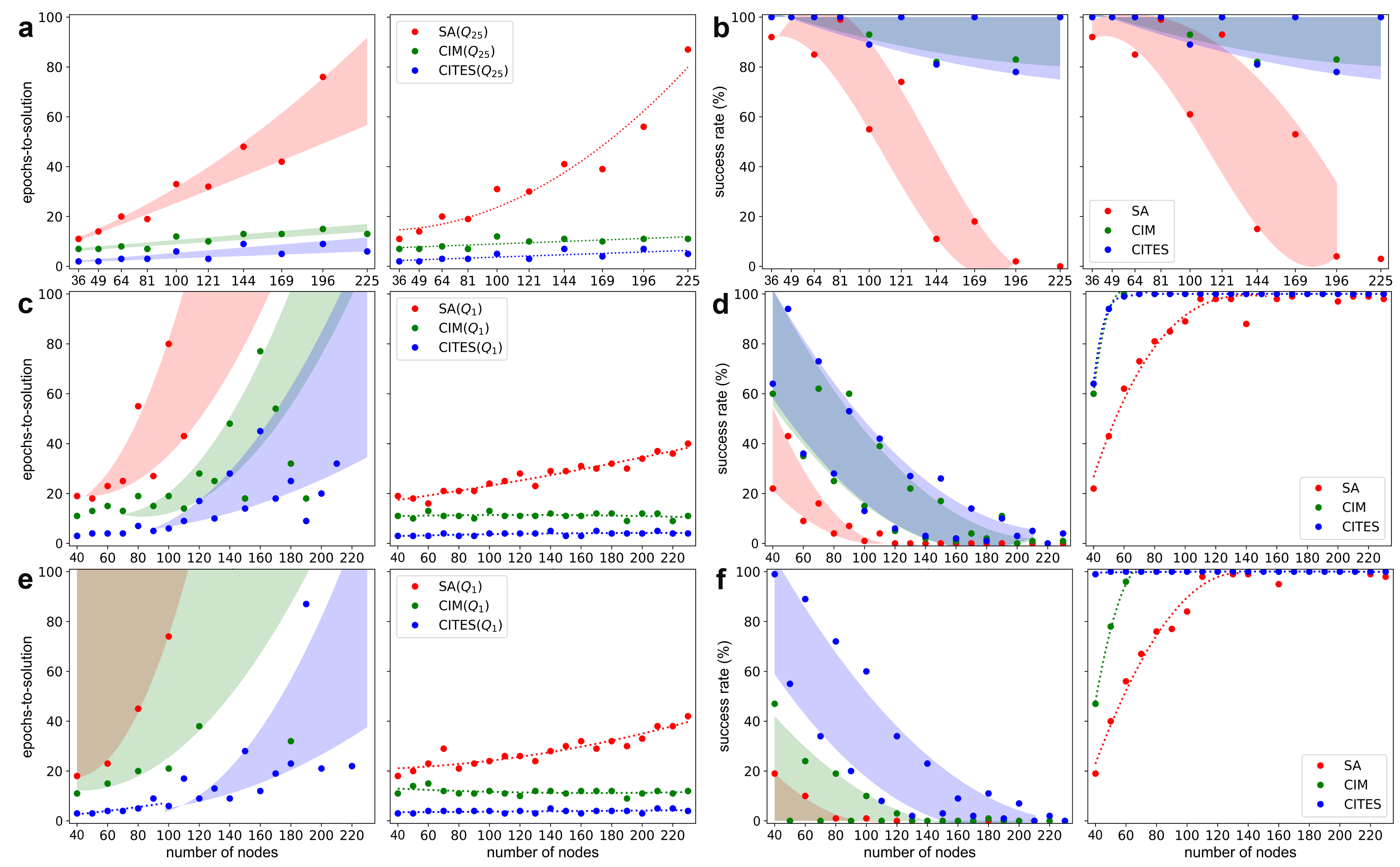}
\caption{\textbf{Benchmark results on square lattice graphs, circular ladder graphs and Mobius ladder graphs:} 
\textbf{a)} Epochs-to-solution of the first quartile ($Q_{25}$) instance, and \textbf{b)} success rate on square lattice graphs with different graph size. The targets of left sub figures are the exact solutions and of right figure are the approximate solutions given by SDP. (parameter used: $\alpha=0.25, \beta=0.29$). 
\textbf{c,d)} are evaluate on circular ladder graphs, and \textbf{e,f)} are evaluate on Mobius ladder graphs ($\alpha=0.07, \beta=0.39$), where epochs-to-solution is evaluate on the first percentile ($Q_1$) instance since the success rates are lower on harder instances.}
\label{fig:benchmark}
\end{figure}

Due to their parallelism, \ac{CITS} and \ac{CIM} simulation demonstrate strong potential in solving 10-by-10 square lattice graphs (Fig.~\ref{fig:10by10lattice}a), which are regarded as easy instances since the Ising spins do not compete with each other. 
As a result, the graphs have only two naive solutions (alternative arranged, $S=(s_1,s_2,...,s_n)$ and $(-s_1,-s_2,...-s_n)$) that are regular graphs.
However, if the side length of the square lattices are odd, the adjacent Ising spins may compete with each other leading to disorder patterns different from Fig.~\ref{fig:10by10lattice}b.
These graphs are known as frustrated graphs, whose Hamiltonian usually has more than two ground states.
Square lattice graphs with side length ranging from 6 to 15 are studied in this section.
The left figure in Fig.~\ref{fig:benchmark}a shows the epochs-to-solution toward the global minima of the first quartile's instances, where they statistically indicate how fast the heuristic algorithms can escape the local energy minima.
We observe that regular graphs usually require more epochs to reach the ground state (upper bounds of the filled area) than frustrated graphs (lower bounds of the filled area).
The epochs-to-solution of \ac{SA} scales as a polynomial function of the number of nodes and the fitting curve corresponding to the regular graphs has a larger higher order coefficient because they only have two global energy minima while the frustrated graphs have multiple energy minima.
For \ac{CIM} and \ac{CITS}, the epochs-to-solution scales linearly with the number of nodes, and the difference between fitting curves of regular graphs and frustrated graph is insignificant. 
They also benefit from the continuous number representation of Ising spins.
\ac{SA} suffers from polynomial scaling due to the binary representation of the Ising spin, which limits the number of spin flips per epoch.
To gain more intuition on the slower convergence when using \ac{SA} as compared to the two other algorithms, let us consider a derandomized and serialized version: Hopfield neural network (HNN)\cite{hopfield1982neural}, which maximally switches one spin at one time.
In the best-case scenario, it will require $2n^2$ time flips to solve a $2n\times 2n$ square lattice.
Notably, \ac{CITS} usually have a lower zero-order coefficient in the fitting curves since a deeper coherent Ising tree tends to make decision based on future time step. 
However, the depth and breadth is only 2 in this case and the benefit of shorter ground state search time diminishes as the number of nodes is increased.
Thus far, our results show that, \ac{CITS} has 2.52x/6.38x speed up compared to \ac{CIM}/SA to find the ground states of the square lattice graphs.
The left figure in Fig.~\ref{fig:benchmark}b shows the success rates of each heuristic algorithm within 100 epochs, revealing how likely that the annealing algorithms get trapped in the local energy minima and return sub-optimal solutions.
The success rates of \ac{CIM} and \ac{CITS} are similar, achieving almost 100\% on the frustrated graphs and decrease logarithmically on the regular graphs, while remaining above 80\%.
Theoretically, \ac{SA} is always expected to find the ground state in infinite time scale and the success rates is closely related to the epochs-to-solution.
SA can achieve 100\% success rate on square lattice graphs with side length 7 or 9. 
However, the success rate decreases polynomially on the frustrated graph. 
Similar to the epochs-to-solution, the success rates on regular graphs are also lower than the frustrated graphs. 

In many cases, a near-optimal solution is acceptable when the time to find the exact solution is incredibly large.
Thus, we also benchmarked these three annealing algorithms to the approximate solutions given by SDP, as shown in the right figures in Fig.~\ref{fig:benchmark}a,b.
SDP is able to achieve exact solution for square lattice with side length of lower than 10, and can only achieve 0.933- to 0.982-approximation for larger graphs.
For \ac{CIM} and \ac{CITS}, the probability of finding near-optimal solutions are similar to those for finding exact solutions of the square lattice, which reveal that they are unlikely to be trapped in local minima in square lattice graphs. 
In this case, \ac{CITS} achieves 2.55x speed up over \ac{CIM} to finding the approximate solutions.
Meanwhile, \ac{SA} show faster convergence speed to find approximate solutions compared to the exact solutions, especially for the regular graphs.
However, it is still 6.38x slower than \ac{CITS}.
Since the performance of \ac{SA} is limited by the speed, relaxed targets are more easily obtained at fewer epochs-to-solution.

In this section, we also benchmark \ac{CITS} on other representative MAX-CUT instances, circular and Mobius ladder graphs with number of nodes ranging from 40 to 230, where some of them are regular graphs and some are frustrated graphs. 
The graph structure of the circular ladder consists of two concentric \emph{n}-cycles and each pair of nodes are connected to each other and the adjacent nodes.
Fig.~\ref{fig:10by10lattice}a) shows that it can be considered as a special case of rectangular lattice with one dimension only have side length of 2 and the other dimension is periodic. 
In the left of Fig.~\ref{fig:benchmark}d, it is clear that among the circular ladder graphs with different number of nodes, \ac{CIM} and \ac{CITS} can only achieve success rate above 75\% on 60 node instance.
Moreover, as the number of nodes increases, the success probabilities will further decrease.
However, \ac{CITS} still has 0.15\%/11.50\% improvement to finding exact solutions compared to \ac{CIM}/SA.
So we adopt the first percentile instead of the twenty-fifth percentile results for evaluating the speed of the algorithms.
The results are shown in the left of the Fig.~\ref{fig:benchmark}c.
Compared to the results in Fig.~\ref{fig:benchmark}a,b), circular ladders can be considered as a more difficult graph topology for MAX-CUT problems, where the fitting curve of epochs-to-solution and success rates of \ac{CIM} and \ac{CITS} both shows exponential scaling.
On average, \ac{CITS} can have 1.97$\times$/2.71$\times$ speed up of finding exact solutions on circular ladder graphs in terms of epoch-to-solution.

The approximate solutions given by SDP are far from the exact solutions, especially when the graph size is large. 
Therefore, these targets become easier to achieve for the annealing algorithms.
Morover, SDP performs worse in difficult instance.
On the right of Fig.~\ref{fig:benchmark}c,d), the curves corresponding to regular graphs and frustrated graphs are fitted jointly.
Interestingly, in Fig.~\ref{fig:benchmark}c, the epochs-to-solution of \ac{CIM} and \ac{CITS} are nearly constant whereas \ac{SA} shows linear dependence on the number of nodes.
The right figure in Fig.~\ref{fig:benchmark}d shows that the success rates are better because SDP performs worse on larger size graph and gives a relaxed target. 
When the number of nodes is above 60, \ac{CIM} and \ac{CITS} can achieve around 100\% success rate to finding approximate solutions whereas \ac{SA} has lower success rate limited by the speed.
On average, \ac{CITS} can have 3.02$\times$/7.21$\times$ speed up as compared to \ac{CIM}/SA and 2.40\%/19.90\% improvement on success rate to finding approximate solutions.

By twisting the blue dash lines in the circular ladder in Fig.~\ref{fig:10by10lattice}a, the graph becomes a Mobius ladder, which is a cubic graph with all the adjacent and opposite nodes connected.
The number of vertices divided by 2 is odd, the Hamiltonian of cutting the Mobius ladder will be minimized when the Ising spins are in the alternate arrangement of up and down along the ring.
When the number of vertices divided by 2 is even instead, all the Ising spins are competing with the corresponding opposite spin in alternate arrangement spin configuration.
The result in Fig.~\ref{fig:benchmark}e,f reveals that, the Mobius ladder graphs are harder than the previous mentioned graphs.
However, on average \ac{CITS} can have 3.12$\times$/7.45$\times$ speed up as compared to \ac{CIM}/SA to finding approximate solutions and 3.42$\times$/8.27$\times$ to finding exact solutions on the Mobius ladder graph.
For the success rate, \ac{CITS} has 3.90\%/14.60\% improvement to finding approximate solutions and 21.35\%/25.00\% improvement to finding exact solutions.

\section*{Discussion}

We have proposed a heuristic search algorithms for Ising formulated combinatorial optimization problem, which we call \ac{CITS}.
This algorithm is inspired by high-level idea of MCTS, which obtains a solution by exploring and exploiting the search space in the feasible regions.
However, for NP-optimization problems, the size of the feasible regions scale exponentially with the size of the problem.
To address this issue, we propose coherent Ising tree search, which is a (recursive-)depth limited search scheme, to find the most potential feasible solutions determined by primal heuristic \ac{SA}.
Our algorithm is also inspired by the coherent Ising machine, which is a quantum-inspired algorithm with the measurement-feedback scheme. 
Since we observe and mathematically formulate bistability of parametric oscillator with spin flip, we proposed two search schemes.
The first scheme explores (using the primal heuristic \ac{CIM}) every feasible region while expanding the Ising tree.
The other scheme is a naive scheme that the same as the first scheme except that the exploration step is removed. 
This reduces the computational complexity and was shown to be more computationally efficient in the section discussing the ablation study.
Our results reveal that the advantage of \ac{CITS} is due mainly to the breadth and the depth of tree instead of the exploring every local search spaces. 
To benchmark the performance of \ac{CITS} as a general solver on Ising formulated problems, we evaluate its performance on MAX-CUT problems, where the instances have nodes ranging from 36 to 230.
\ac{CITS} has improvement in both epochs-to-solution and success rate as compared to the simulated poor man's \ac{CIM} on both square lattice graphs, circular ladder graphs and Mobius ladder graphs, where \ac{CITS} has maximally up to 3.42$\times$ speed up and 21.35\% higher success rate.
The implementation of \ac{CITS} that interface with physical \ac{CIM} is left for future work.
We note that when physically implementing \ac{CIM} with the measurement-feedback scheme, the bottleneck of the time-to-solution is mainly the communication between FPGA and the oscillators.
Thus, it is reasonable to expect that the speed up in epochs-to-solution is similar to the speed up in computation time.

\section*{Methods}

\subsection*{Problem Mapping and Baseline}

It is well known that the MAX-CUT problems belong to the NP-hard class\cite{karp1972reducibility}, which means any NP problem can be mapped to the MAX-CUT problem in polynomial time. 
The description of the MAX-CUT problem is as follows. Given an undirected graph $G=\left(V,E\right)$, partition $G$ into two complementary graphs, $G_1$ and $G_2$, such that the number of edges between $G_1$ and $G_2$ is maximized. 
The objective function can be written as:

\begin{equation}
\text{CUT}\left(s_1,s_2,\ ...,s_n\right)=\sum_{\left(i,j\right)\in E}\frac{1-s_is_j}{2}
\label{eq:maxcut}
\end{equation}
Comparing Eq.~\eqref{eq:classical_H} and Eq.~\eqref{eq:maxcut}, we observe that maximizing $\text{CUT}\left(s_1,s_2,\ ...,s_n\right)$ is equivalent to minimizing $H\left(s_1,s_2,\ ...,s_n\right)$, when $J_{ij}=1,\ \forall\left(i,j\right)\in E$. 

For most MAX-CUT problems, there is no guarantee that exact solutions can be found in polynomial time. 
Generally, an acceptable solution (one that outperforms some baseline) found within an acceptable time (in polynomial time) is sufficient.
In this work, Goemans-Williamson Semidefinite programming (GW-SDP), a 0.879-approximation algorithm for the MAX-CUT problems\cite{goemans1995improved}, is chosen as the baseline algorithm for generating the targets with which to evaluate the efficiency of \ac{CITS} or its primal heuristics. 
SDP is known as an approximation algorithm which relaxes integer linear programming problem in Eq.~\eqref{eq:maxcut} to:

\begin{equation}
\text{CUT}\left(\sigma_{11},\sigma_{12},\ ...,\sigma_{nn}\right)=\sum_{\left(i,j\right)\in E}\frac{1-\sigma_{ij}}{2}
\label{eq:sdp}
\end{equation}
$\sigma_{ij}=s_is_j\in\left\{-1,1\right\}$, represents whether two vertices are in the same subgraph or not. The constraints of SDP are: $\sigma_{ii} =  1$, $\sigma_{ij} = \sigma_{ji}$, and $\Sigma = \left[\sigma_{ij}\right]$ is positive semidefinite.
As a result, the MAX-CUT problems are transformed into a special case of convex programming, in which the \emph{cvx} solver\cite{becker2011templates} can efficiently maximize the objective function in Eq.~\eqref{eq:sdp} with the aforementioned constraints.
After transformation, a Cholesky factorization may be performed on $\Sigma=S^{T}S$.
A projection of $S\approx\text{sgn}\left[\Sigma B\right]$ via random rounding may then be utilized to approximate the solution.
$B$ is an $n\times{}m$ random matrix, where $m$ columns represent $m$ random planes to be projected on to generate approximate solutions.
Finally, among all approximate solution, the one giving the largest number of cuts will be selected.

\subsection*{Primal heuristic 1: Parallel \ac{SA}}

The stochastic spin update scheme in \ac{SA} is inspired by thermal annealing, where the probability of the spin-flip is determined by quasi-equilibrium distribution based on the Metropolis-Hasting\cite{metropolis1953equation} algorithm:

\begin{equation}
p_i=\frac{1}{Z}\exp\left(-\frac{\Delta H_i}{k_BT}\right)
\label{eq:prob}
\end{equation}
$p_i$ is the flipping probability of the $i$-th spin, $Z$ is a normalization factor, $k_B$ is the Boltzmann constant, $T$ is the annealing temperature parameter, and $\Delta H_i$ is the energy difference due to flipping the $i$-th spin. 
Note that the ergodic search of all possible spin configurations requires a $1\times2^n$ probability transition matrix. 
As a result, following the exact Boltzmann distribution becomes difficult. Instead, a quasi-equilibrium distribution only concerning the flipping probability of each spin is utilized. 
In this scenario, the normalization term is approximated as $Z = \sum_{i=1}^n \exp\left(-\frac{\Delta H_i}{k_BT}\right)$. Typically, the energy change in flipping the $i^{th}$ spin is:

\begin{equation}
\Delta H_i = -2s_i\sum_i^nJ_{ij}s_j
\label{eq:Delta}
\end{equation}
To speed up the algorithm, we adopt a synchronous update by approximating Eq.~\eqref{eq:prob} to $p_i \gets T^\ast p_i/\sum_{i}^{n}p_i$, where $k_BT=1$. 
Note that the quasi-temperature parameter, $T^\ast$, is initialized as 1 and subject to an temperature scheduling scheme\cite{mills2020finding} as follows. Increase the temperature significantly in the first few epochs to provide thermal energy to escape the minimum. Thereafter, the temperature is aggressively decreased in the next few epochs before slowing the temperature decrease to a gradual decay.

\begin{equation}
\Delta T^\ast(t+1) =
\begin{cases}
1.05T^\ast(t), & t\le\frac{N_{epochs}}{4}\\
0.95T^\ast(t), & \frac{N_{epochs}}{4}<t\le\frac{N_{epochs}}{2}\\
0.99T^\ast(t), & t>\frac{N_{epochs}}{2}\\
\end{cases}
\label{eq:temp}
\end{equation}

\subsection*{Primal heuristic 2: Simulated Poorman's \ac{CIM}}

Another primal heuristic is \ac{CIM}, which encodes the coherent Ising spin state as the phases of degenerate oscillators.
With mutual injections of the signals between each oscillator, the \ac{CIM} will oscillate in one of the approximate ground states, thereby giving near-optimal solutions to the combinatorial optimization problem.
The mutual injections are usually realized by the network of delay lines\cite{marandi2014network}, or approximated by measurement-feedback scheme\cite{mcmahon2016fully} where the latter is discretized by the Euler method. 
The simulation in the present work follows the poorman's \ac{CIM}\cite{bohm2019poor}, where the time evolution of the $i$-th spin at time step $t$ is:

\begin{equation}
x_i\left[t+1\right] = \cos^2\left(f_i\left[t\right]-\frac{\pi}{4}+\xi_i\left[t\right]\right) - \frac{1}{2}\\
\label{eq:x}
\end{equation}
$x_i\left[t\right]$ is the measurement of each Ising spin at time step $t$, and $\xi_i\left[t\right]$ is the diffusion introduced by modelled using Gaussian white noise.
Note that the noise for poor man's \ac{CIM} is inside the trigonometric function whereas it is outside for \ac{CITS}.
For \ac{CIM}, the noise is modelled as $\xi\sim\mathbb{N}(0,10^{-2})$, and the same level of noise is applied at the beginning for random initialization and removed at the later time steps for a deterministic convergence.
$f_i\left[t\right]$ is the feedback term injected back to each of the Ising spins:

\begin{equation}
f_i\left[t\right] = \alpha x_i\left[t\right] +\beta\sum_{j}^{n}{J_{ij}x_j\left[t\right]}
\label{eq:f}
\end{equation}
The feedback gain, $\alpha$, and coupling gain, $\beta$, remain the same for both \ac{CIM} and \ac{CITS}, but should be chosen carefully to ensure bifurcations of Ising spins. After trial-and-error, we select $\alpha/\beta$ for the 2D square lattices as $0.25/0.29$, and as $0.07/0.39$ for circular ladder, Mobius ladders. 

\subsection*{High-level strategy: \ac{CITS} Algorithm}

\begin{figure}[ht]
\centering
\includegraphics[width=\linewidth]{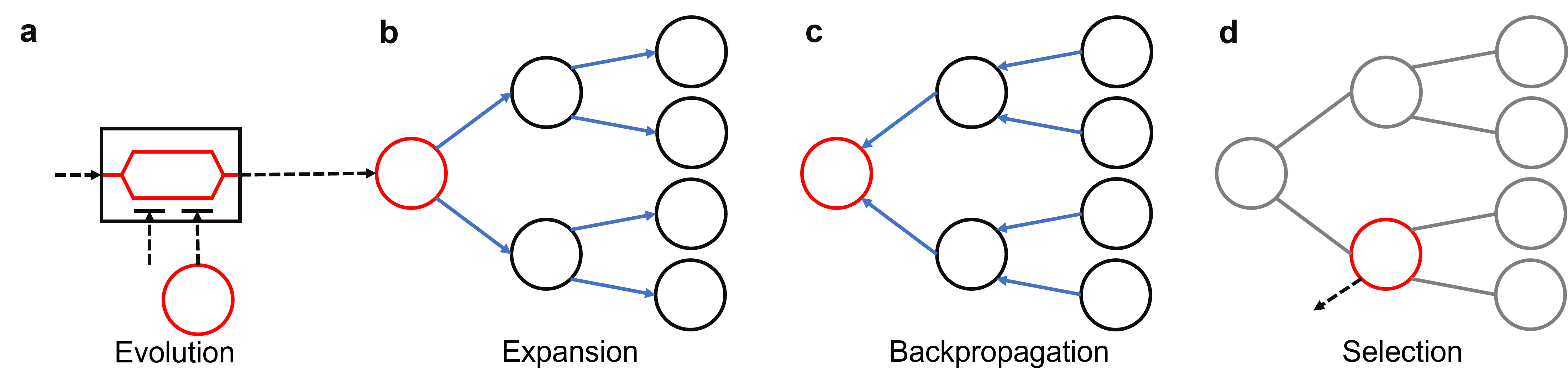}
\caption{\textbf{An illustrative workflow of \ac{CITS} with tree depth of $d=2$ and breadth of $b=2$}: 
\textbf{a)} Ising spin evolution in simulated (or physical) \ac{CIM}, the reduced dynamics of the oscillators tends to approach lower Hamiltonian.
\textbf{b)} Expansion of coherent Ising tree is based on the flipping probability given by the primal heuristic \ac{SA}, where the most potential flipping will be expanded as a child nodes. 
\textbf{c)} \ac{CITS} computes the rewards $R$ of each node based on the Ising Hamiltonian of the corresponding spin configuration. And the return $Q$ of the child nodes will backpropagte to the parent nodes for computing their return.
\textbf{d)} \ac{CITS} select the nodes with highest return successively, until reaching the leaf node, or all the child nodes have negative return.
}
\label{fig:CITES}
\end{figure}

The workflow of \ac{CITS} (Fig.~\ref{fig:CITES}) consists of four main steps.

\begin{enumerate}
    \item Evolution: As shown in Fig.~\ref{fig:CITES}a, the Ising formulated graph is encoded onto the simulated \ac{CIM} (or its physical implementation).
    The parametric oscillators are then allowed to interact with each other while evolving.
    The obtained (measured) result from the \ac{CIM} is then used to initialize the root node of the coherent Ising tree.
    \item Expansion (and Exploration): For all nodes in the current layer, compute the switching probability, $p_i$, according Eq. \eqref{eq:prob} (the primal heuristic \ac{SA}).
    Thereafter, create $b$ child nodes based on top-$b$ probabilities, which will be further given to the created child nodes as the prior, $p$. 
    Note that the $i$-th spin will take the value of the opposite number, \emph{i.e.}, $X_i^{(k)} = \left(x_1, x_2, ..., -x_i, ..., x_n\right)$. 
    Afterward, the child nodes will explore the search space using the primal heuristic \ac{CIM} (Eq.~\eqref{eq:reduced_dynamic}).
    A naive version to reduce the computational complexity is obtained by removing the exploration step and using the binary Ising spin to represent the Ising spins.
    The expected reward of each node in both schemes may be computed by:
    \begin{equation}
        R_i^{(k)}=H\left(\text{sgn}[X_i^{(k)}]\right)-H\left(\text{sgn}[X^{(0)}]\right)
        \label{eq:leaf}
    \end{equation}
    where $X_i^{(k)}$ is the spin configuration corresponding to the $i$-th node in the $d$-th layer. 
    This step is repeated for these child nodes if they have not reached the tree depth $d$;
    \item Backpropagation: Starting from the nodes at deepest layer, \ac{CITS} samples the return, $Q_j^{(k)}$, of each node.
    The return is successively backpropagated to their parent nodes (following the blue arrow lines in Fig.~\ref{fig:CITES}c) until the root node is reached.
    The return of the $j$-th node in $k$-th layer is computed using the prior $p_i^{(k+1)}$ and the expected rewards of the child nodes as:
    \begin{equation}
        Q_j^{(k)}=p_i^{(k+1)}R_i^{(k)}+\sum_{i\in C_j^{(k)}}{p_i^{(k+1)}Q_i^{(k+1)}}
        \label{eq:reward}
    \end{equation}
    where $C_j^{(k)}$ is the set containing all child nodes of the $j$-th node in $k$-th layer;
    \item Selection: Starting from the root node to a chosen layer, select the child node with the highest return $Q_j^{(k)}$ successively.
    If the current node does not have any child node with positive return, stop selection and provide the spin configuration for evolution.
    The selected child node is given to the simulated (or physical) \ac{CIM} for the evolution in the next time step.
\end{enumerate}

\bibliography{reference}

\begin{thebibliography}{10}
\urlstyle{rm}
\expandafter\ifx\csname url\endcsname\relax
  \def\url#1{\texttt{#1}}\fi
\expandafter\ifx\csname urlprefix\endcsname\relax\def\urlprefix{URL }\fi
\expandafter\ifx\csname doiprefix\endcsname\relax\def\doiprefix{DOI: }\fi
\providecommand{\bibinfo}[2]{#2}
\providecommand{\eprint}[2][]{\url{#2}}

\bibitem{anand2012customized}
\bibinfo{author}{Anand, S.}, \bibinfo{author}{Saravanasankar, S.} \&
  \bibinfo{author}{Subbaraj, P.}
\newblock \bibinfo{journal}{\bibinfo{title}{Customized simulated annealing
  based decision algorithms for combinatorial optimization in vlsi
  floorplanning problem}}.
\newblock {\emph{\JournalTitle{Computational Optimization and Applications}}}
  \textbf{\bibinfo{volume}{52}}, \bibinfo{pages}{667--689}
  (\bibinfo{year}{2012}).

\bibitem{matsubara2020digital}
\bibinfo{author}{Matsubara, S.} \emph{et~al.}
\newblock \bibinfo{title}{Digital annealer for high-speed solving of
  combinatorial optimization problems and its applications}.
\newblock In \emph{\bibinfo{booktitle}{2020 25th Asia and South Pacific Design
  Automation Conference (ASP-DAC)}}, \bibinfo{pages}{667--672}
  (\bibinfo{organization}{IEEE}, \bibinfo{year}{2020}).

\bibitem{tanahashi2019application}
\bibinfo{author}{Tanahashi, K.}, \bibinfo{author}{Takayanagi, S.},
  \bibinfo{author}{Motohashi, T.} \& \bibinfo{author}{Tanaka, S.}
\newblock \bibinfo{journal}{\bibinfo{title}{Application of ising machines and a
  software development for ising machines}}.
\newblock {\emph{\JournalTitle{Journal of the Physical Society of Japan}}}
  \textbf{\bibinfo{volume}{88}}, \bibinfo{pages}{061010}
  (\bibinfo{year}{2019}).

\bibitem{lucas2014ising}
\bibinfo{author}{Lucas, A.}
\newblock \bibinfo{journal}{\bibinfo{title}{Ising formulations of many np
  problems}}.
\newblock {\emph{\JournalTitle{Frontiers in physics}}}
  \textbf{\bibinfo{volume}{2}}, \bibinfo{pages}{5} (\bibinfo{year}{2014}).

\bibitem{johnson2011quantum}
\bibinfo{author}{Johnson, M.~W.} \emph{et~al.}
\newblock \bibinfo{journal}{\bibinfo{title}{Quantum annealing with manufactured
  spins}}.
\newblock {\emph{\JournalTitle{Nature}}} \textbf{\bibinfo{volume}{473}},
  \bibinfo{pages}{194--198} (\bibinfo{year}{2011}).

\bibitem{kirkpatrick1983optimization}
\bibinfo{author}{Kirkpatrick, S.}, \bibinfo{author}{Gelatt, C.~D.} \&
  \bibinfo{author}{Vecchi, M.~P.}
\newblock \bibinfo{journal}{\bibinfo{title}{Optimization by simulated
  annealing}}.
\newblock {\emph{\JournalTitle{science}}} \textbf{\bibinfo{volume}{220}},
  \bibinfo{pages}{671--680} (\bibinfo{year}{1983}).

\bibitem{mitra1986convergence}
\bibinfo{author}{Mitra, D.}, \bibinfo{author}{Romeo, F.} \&
  \bibinfo{author}{Sangiovanni-Vincentelli, A.}
\newblock \bibinfo{journal}{\bibinfo{title}{Convergence and finite-time
  behavior of simulated annealing}}.
\newblock {\emph{\JournalTitle{Advances in applied probability}}}
  \textbf{\bibinfo{volume}{18}}, \bibinfo{pages}{747--771}
  (\bibinfo{year}{1986}).

\bibitem{mu202120x28}
\bibinfo{author}{Mu, J.}, \bibinfo{author}{Su, Y.} \& \bibinfo{author}{Kim, B.}
\newblock \bibinfo{title}{A 20x28 spins hybrid in-memory annealing computer
  featuring voltage-mode analog spin operator for solving combinatorial
  optimization problems}.
\newblock In \emph{\bibinfo{booktitle}{2021 Symposium on VLSI Technology}},
  \bibinfo{pages}{1--2} (\bibinfo{organization}{IEEE}, \bibinfo{year}{2021}).

\bibitem{sutton2017intrinsic}
\bibinfo{author}{Sutton, B.}, \bibinfo{author}{Camsari, K.~Y.},
  \bibinfo{author}{Behin-Aein, B.} \& \bibinfo{author}{Datta, S.}
\newblock \bibinfo{journal}{\bibinfo{title}{Intrinsic optimization using
  stochastic nanomagnets}}.
\newblock {\emph{\JournalTitle{Scientific reports}}}
  \textbf{\bibinfo{volume}{7}}, \bibinfo{pages}{1--9} (\bibinfo{year}{2017}).

\bibitem{shim2017stochastic}
\bibinfo{author}{Shim, Y.}, \bibinfo{author}{Jaiswal, A.} \&
  \bibinfo{author}{Roy, K.}
\newblock \bibinfo{title}{Stochastic switching of she-mtj as a natural annealer
  for efficient combinatorial optimization}.
\newblock In \emph{\bibinfo{booktitle}{2017 IEEE International Conference on
  Computer Design (ICCD)}}, \bibinfo{pages}{605--608}
  (\bibinfo{organization}{IEEE}, \bibinfo{year}{2017}).

\bibitem{mondal2020ising}
\bibinfo{author}{Mondal, A.} \& \bibinfo{author}{Srivastava, A.}
\newblock \bibinfo{journal}{\bibinfo{title}{Ising-fpga: A spintronics-based
  reconfigurable ising model solver}}.
\newblock {\emph{\JournalTitle{ACM Transactions on Design Automation of
  Electronic Systems (TODAES)}}} \textbf{\bibinfo{volume}{26}},
  \bibinfo{pages}{1--27} (\bibinfo{year}{2020}).

\bibitem{andrawis2020antiferroelectric}
\bibinfo{author}{Andrawis, R.} \& \bibinfo{author}{Roy, K.}
\newblock \bibinfo{journal}{\bibinfo{title}{Antiferroelectric tunnel junctions
  as energy-efficient coupled oscillators: Modeling, analysis, and application
  to solving combinatorial optimization problems}}.
\newblock {\emph{\JournalTitle{IEEE Transactions on Electron Devices}}}
  \textbf{\bibinfo{volume}{67}}, \bibinfo{pages}{2974--2980}
  (\bibinfo{year}{2020}).

\bibitem{hibat2021variational}
\bibinfo{author}{Hibat-Allah, M.}, \bibinfo{author}{Inack, E.~M.},
  \bibinfo{author}{Wiersema, R.}, \bibinfo{author}{Melko, R.~G.} \&
  \bibinfo{author}{Carrasquilla, J.}
\newblock \bibinfo{journal}{\bibinfo{title}{Variational neural annealing}}.
\newblock {\emph{\JournalTitle{Nature Machine Intelligence}}}
  \textbf{\bibinfo{volume}{3}}, \bibinfo{pages}{952–961}
  (\bibinfo{year}{2021}).

\bibitem{marandi2014network}
\bibinfo{author}{Marandi, A.}, \bibinfo{author}{Wang, Z.},
  \bibinfo{author}{Takata, K.}, \bibinfo{author}{Byer, R.~L.} \&
  \bibinfo{author}{Yamamoto, Y.}
\newblock \bibinfo{journal}{\bibinfo{title}{Network of time-multiplexed optical
  parametric oscillators as a coherent ising machine}}.
\newblock {\emph{\JournalTitle{Nature Photonics}}}
  \textbf{\bibinfo{volume}{8}}, \bibinfo{pages}{937--942}
  (\bibinfo{year}{2014}).

\bibitem{mcmahon2016fully}
\bibinfo{author}{McMahon, P.~L.} \emph{et~al.}
\newblock \bibinfo{journal}{\bibinfo{title}{A fully programmable 100-spin
  coherent ising machine with all-to-all connections}}.
\newblock {\emph{\JournalTitle{Science}}} \textbf{\bibinfo{volume}{354}},
  \bibinfo{pages}{614--617} (\bibinfo{year}{2016}).

\bibitem{bohm2019poor}
\bibinfo{author}{B{\"o}hm, F.}, \bibinfo{author}{Verschaffelt, G.} \&
  \bibinfo{author}{Van~der Sande, G.}
\newblock \bibinfo{journal}{\bibinfo{title}{A poor man’s coherent ising
  machine based on opto-electronic feedback systems for solving optimization
  problems}}.
\newblock {\emph{\JournalTitle{Nature communications}}}
  \textbf{\bibinfo{volume}{10}}, \bibinfo{pages}{1--9} (\bibinfo{year}{2019}).

\bibitem{coulom2006efficient}
\bibinfo{author}{Coulom, R.}
\newblock \bibinfo{title}{Efficient selectivity and backup operators in
  monte-carlo tree search}.
\newblock In \emph{\bibinfo{booktitle}{International conference on computers
  and games}}, \bibinfo{pages}{72--83} (\bibinfo{organization}{Springer},
  \bibinfo{year}{2006}).

\bibitem{sherrington1975solvable}
\bibinfo{author}{Sherrington, D.} \& \bibinfo{author}{Kirkpatrick, S.}
\newblock \bibinfo{journal}{\bibinfo{title}{Solvable model of a spin-glass}}.
\newblock {\emph{\JournalTitle{Physical review letters}}}
  \textbf{\bibinfo{volume}{35}}, \bibinfo{pages}{1792} (\bibinfo{year}{1975}).

\bibitem{ising1925beitrag}
\bibinfo{author}{Ising, E.}
\newblock \bibinfo{journal}{\bibinfo{title}{Beitrag zur theorie des
  ferromagnetismus}}.
\newblock {\emph{\JournalTitle{Zeitschrift f{\"u}r Physik}}}
  \textbf{\bibinfo{volume}{31}}, \bibinfo{pages}{253--258}
  (\bibinfo{year}{1925}).

\bibitem{wang2019oim}
\bibinfo{author}{Wang, T.} \& \bibinfo{author}{Roychowdhury, J.}
\newblock \bibinfo{title}{Oim: Oscillator-based ising machines for solving
  combinatorial optimisation problems}.
\newblock In \emph{\bibinfo{booktitle}{International Conference on
  Unconventional Computation and Natural Computation}},
  \bibinfo{pages}{232--256} (\bibinfo{organization}{Springer},
  \bibinfo{year}{2019}).

\bibitem{wang2013coherent}
\bibinfo{author}{Wang, Z.}, \bibinfo{author}{Marandi, A.},
  \bibinfo{author}{Wen, K.}, \bibinfo{author}{Byer, R.~L.} \&
  \bibinfo{author}{Yamamoto, Y.}
\newblock \bibinfo{journal}{\bibinfo{title}{Coherent ising machine based on
  degenerate optical parametric oscillators}}.
\newblock {\emph{\JournalTitle{Physical Review A}}}
  \textbf{\bibinfo{volume}{88}}, \bibinfo{pages}{063853}
  (\bibinfo{year}{2013}).

\bibitem{goto2019combinatorial}
\bibinfo{author}{Goto, H.}, \bibinfo{author}{Tatsumura, K.} \&
  \bibinfo{author}{Dixon, A.~R.}
\newblock \bibinfo{journal}{\bibinfo{title}{Combinatorial optimization by
  simulating adiabatic bifurcations in nonlinear hamiltonian systems}}.
\newblock {\emph{\JournalTitle{Science advances}}}
  \textbf{\bibinfo{volume}{5}}, \bibinfo{pages}{eaav2372}
  (\bibinfo{year}{2019}).

\bibitem{hopfield1982neural}
\bibinfo{author}{Hopfield, J.~J.}
\newblock \bibinfo{journal}{\bibinfo{title}{Neural networks and physical
  systems with emergent collective computational abilities}}.
\newblock {\emph{\JournalTitle{Proceedings of the national academy of
  sciences}}} \textbf{\bibinfo{volume}{79}}, \bibinfo{pages}{2554--2558}
  (\bibinfo{year}{1982}).

\bibitem{karp1972reducibility}
\bibinfo{author}{Karp, R.~M.}
\newblock \bibinfo{title}{Reducibility among combinatorial problems}.
\newblock In \emph{\bibinfo{booktitle}{Complexity of computer computations}},
  \bibinfo{pages}{85--103} (\bibinfo{publisher}{Springer},
  \bibinfo{year}{1972}).

\bibitem{goemans1995improved}
\bibinfo{author}{Goemans, M.~X.} \& \bibinfo{author}{Williamson, D.~P.}
\newblock \bibinfo{journal}{\bibinfo{title}{Improved approximation algorithms
  for maximum cut and satisfiability problems using semidefinite programming}}.
\newblock {\emph{\JournalTitle{Journal of the ACM (JACM)}}}
  \textbf{\bibinfo{volume}{42}}, \bibinfo{pages}{1115--1145}
  (\bibinfo{year}{1995}).

\bibitem{becker2011templates}
\bibinfo{author}{Becker, S.~R.}, \bibinfo{author}{Cand{\`e}s, E.~J.} \&
  \bibinfo{author}{Grant, M.~C.}
\newblock \bibinfo{journal}{\bibinfo{title}{Templates for convex cone problems
  with applications to sparse signal recovery}}.
\newblock {\emph{\JournalTitle{Mathematical programming computation}}}
  \textbf{\bibinfo{volume}{3}}, \bibinfo{pages}{165} (\bibinfo{year}{2011}).

\bibitem{metropolis1953equation}
\bibinfo{author}{Metropolis, N.}, \bibinfo{author}{Rosenbluth, A.~W.},
  \bibinfo{author}{Rosenbluth, M.~N.}, \bibinfo{author}{Teller, A.~H.} \&
  \bibinfo{author}{Teller, E.}
\newblock \bibinfo{journal}{\bibinfo{title}{Equation of state calculations by
  fast computing machines}}.
\newblock {\emph{\JournalTitle{The journal of chemical physics}}}
  \textbf{\bibinfo{volume}{21}}, \bibinfo{pages}{1087--1092}
  (\bibinfo{year}{1953}).

\bibitem{mills2020finding}
\bibinfo{author}{Mills, K.}, \bibinfo{author}{Ronagh, P.} \&
  \bibinfo{author}{Tamblyn, I.}
\newblock \bibinfo{journal}{\bibinfo{title}{Finding the ground state of spin
  hamiltonians with reinforcement learning}}.
\newblock {\emph{\JournalTitle{Nature Machine Intelligence}}}
  \textbf{\bibinfo{volume}{2}}, \bibinfo{pages}{509--517}
  (\bibinfo{year}{2020}).

\end{thebibliography}

\section*{Data availability}
The data that support the results of this study are available from the corresponding
author upon reasonable request.

\section*{Acknowledgements}

This work is funded in part by the  Agency for Science, Technology and Research (A*STAR), Singapore, under its Programmatic Grant Programme (SpOT-LITE) and by the National Research Foundation, Singapore, under its the Competitive Research Programme (CRP Award No. NRF-CRP24-2020-0003).

\section*{Author contributions}

Y.C. conceived the algorithms and performed the experiments,  Y.C. and D.D. conducted the theoretical analysis, Y.C. and X.F. analysed the results and wrote the manuscript.  All authors reviewed the manuscript. 

\section*{Competing interests}

The authors declare no competing interests.

\section*{Correspondence}

Correspondence and requests for materials should be addressed to X. Fong.

\end{document}